\documentclass[parskip=full,11pt]{article}

\usepackage{amssymb}
\usepackage{amsthm}
\usepackage{amsmath}
\usepackage{amsmath}
\usepackage{float,booktabs}
\usepackage{threeparttable}
\usepackage{float} 
\usepackage{empheq}
\usepackage{graphicx}
\usepackage{hyperref}
\usepackage{natbib}
\usepackage{url}
\usepackage{xcolor}
\usepackage{dsfont}
\usepackage{soul,color,colortbl}
\usepackage{array,multirow}
\usepackage{cancel}
\DeclareMathOperator*{\argmin}{arg\,min} 
\usepackage[top=1in, bottom=1.3in, left=0.875in, right=0.875in]{geometry}
\usepackage[linesnumbered,ruled]{algorithm2e}
\usepackage[labelformat=simple]{subcaption}
\usepackage{bm}
\usepackage[linesnumbered,ruled]{algorithm2e}
\usepackage{caption}
\usepackage{subcaption}

\def\spacingset#1{\renewcommand{\baselinestretch}
{#1}\small\normalsize} \spacingset{1}

\newcolumntype{P}[1]{>{\centering\arraybackslash}p{#1}}
\newcommand*{\myfont}{\fontfamily{lmss}\selectfont}
\DeclareTextFontCommand{\textpython}{\myfont}

\renewcommand{\figurename}{\textbf{Figure}}
\DeclareCaptionLabelFormat{boldnumber}{\bothIfFirst{#1}{~}\textbf{#2}}
\renewcommand{\tablename}{\textbf{Table}}
\title{Enhanced Gradient Boosting for Zero-Inflated Insurance Claims and Comparative Analysis of \texttt{CatBoost}, \texttt{XGBoost}, and \texttt{LightGBM} }

\author{Banghee So \thanks{Department of Mathematics, Towson University, 7800 York Rd, Towson, 21204, MD, USA. Email: \texttt{bso@towson.edu}.}}

\begin{document}

\maketitle

\begin{abstract}
The property and casualty (P\&C) insurance industry faces challenges in developing claim predictive models due to the highly right-skewed distribution of positive claims with excess zeros. To address this, actuarial science researchers have employed ``zero-inflated" models that combine a traditional count model and a binary model. This paper investigates the use of boosting algorithms to process insurance claim data, including zero-inflated telematics data, to construct claim frequency models. Three popular gradient boosting libraries - \texttt{XGBoost}, \texttt{LightGBM}, and \texttt{CatBoost} - are evaluated and compared to determine the most suitable library for training insurance claim data and fitting actuarial frequency models. Through a comprehensive analysis of two distinct datasets, it is determined that \texttt{CatBoost} is the best for developing auto claim frequency models based on predictive performance. Furthermore, we propose a new zero-inflated Poisson boosted tree model, with variation in the assumption about the relationship between inflation probability $p$ and distribution mean $\mu$, and find that it outperforms others depending on the characteristics of the data. This model enables us to take advantage of particular \texttt{CatBoost} tools, which makes it easier and more convenient to investigate the effects and interactions of various risk features on the frequency model when using telematics data.

\vspace{0.8cm}

\noindent \textbf{Keywords}: Auto telematics, Usage-based insurance, Boosted trees, Gradient Boosting, Zero-inflated distribution, Zero-inflated Poisson, \texttt{CatBoost}, \texttt{LightGBM}, \texttt{XGBoost}
\end{abstract}

\newpage

\section{Introduction} \label{sec:intro}

In the property and casualty (P\&C) insurance industry, typical portfolios often exhibit a highly right-skewed distribution for positive claims, with a probability mass at zero in the event of non-occurring claims. It is well known that given these particular characteristics of insurance data, Poisson or negative binomial Generalized Linear Models (GLMs) are commonly used to construct insurance claim frequency models. However, these traditional models face challenges when dealing with insurance claim data where extra dispersion appears due to the number of observed zeros exceeding the number expected under these traditional distribution assumptions. In response to this, the insurance sector has adopted ``zero-inflated" models to handle datasets with excess zero values. Zero-inflated models employ a mixture model strategy that combines two distinct models; a traditional count model, and a binary model, which determines whether a zero is an excess zero (i.e., a zero resulting from a separate process) or a true zero (i.e., a zero resulting from the count model). The examples of this strategy in use in the insurance field include zero-inflated Poisson or negative binomial regression (\citet{yip2005modeling}; \citet{mouatassim2012poisson}; \citet{chen2019subgroup}). These approaches effectively deal with excess zeros in the claim data, thereby offering a more precise understanding and prediction of the claim patterns.

While the Generalized Linear Model (GLM) has been widely applied in actuarial studies (\citet{ayuso2019auto}; \citet{lemaire2016use}), its logarithmic mean structure, constrained to a linear form, can be too inflexible for certain applications, particularly when nonlinearity exists in claim data.  \citet{boucher2017gam}, \citet{henckaerts2018data} and \citet{verbelen2018telem} have expanded the use of GLMs to Generalized Additive Models (GAMs) to more appropriately capture the nonlinear effects of driving behavior features. GLMs and GAMs are powerful; however, they often struggle to identify complex interactions among numerous, highly overlapping risk features. This shortcoming comes primarily from its linear or additive structure, which may not capture the relationships between variables effectively. Furthermore, these models typically require explicit specification of interaction terms, which can be challenging when dealing with a large number of potential interactions or when the nature of these interactions are not predefined. The advent of telematics car driving data, integral to Usage-Based Insurance (UBI) products, has added a new level of complexity to risk modeling in the insurance industry. Telematics data involve information about driving behaviors such as annualized time one the road, braking and acceleration habits, the intensity of left or right turns, and total annual distance driven. These variables can interact with one another or even with traditional variables. For example, the risk associated with driving long distances could be influenced by the age or braking habits of the vehicle. Therefore, modeling interaction effects becomes even more important when telematics data is used. Consequently, recent research has shifted focus towards the use of machine learning techniques to analyze telematics car driving data since they are well-suited to identify and model complex interactions among a high number of variables.

Machine learning in actuarial research has two primary applications. The first of these applications often involves the use of machine learning classifiers. These are tasked with determining the variables that have a significant effect on insurance risk by analyzing auto claim counts. Several studies have successfully implemented these classifiers, including \citet{paefgen2013evaluation}, which used decision trees and artificial neural networks, and \citet{baecke2017value}, which highlighted the importance of telematics data through the application of a random forest classifier. Ensemble learning methods have also proven to be effective in this context, as demonstrated by \citet{bian2018good}. Other algorithms, such as \texttt{XGBoost}, have been recognized for their potency by researchers such as \citet{pesantez2019predicting}, who found them to outperform logistic regression when applied to telematics data. A range of other techniques, including support vector machines, random forests, \texttt{XGBoost}, and artificial neural networks have been compared for their efficacy by \citet{huang2019automobile}. Researchers have even explored deep learning methods, such as \citet{gao2019convolutional}, who trained a deep ConvNet to classify individual trips using variables such as speed and changes in angles. \citet{so2020cost} advanced the field by developing a method, termed SAMM.C2, which combined the SAMME and Ada.C2 algorithms for the analysis of highly imbalanced multiclass telematic data sets.

The second major application of machine learning in actuarial science involves the construction of predictive models for the frequency or severity of insurance claims. This is often accomplished using boosting techniques, which have seen a rise in popularity due to their impressive predictive performance. \citet{guelman2012gradient} pioneered the use of gradient boosting trees with squared-error loss for building frequency and severity models. These techniques have since been expanded upon by researchers like \citet{yang2018insurance}, who applied gradient boosting trees to Tweedie compound Poisson models to predict pure premiums. Other researchers, like \citet{lee2018delta}, \citet{lee2021addressing}, and \citet{henckaerts2021boosting}, continued to innovate and refine these methods, while \citet{hainaut2022response} expanded their application to the response Tweedie loss function for more accurate models. Researchers such as \citet{meng2022actuarial} and \citet{zhou2022tweedie} utilized boosting to enhance prediction accuracy under a zero-inflated assumption. 

Despite progress, a significant gap still exists: The common approach for zero-inflated models usually requires separate training of the models for the inflation probability $p$ and the distribution mean $\mu$. This division presents difficulties for those who want to perform a detailed feature risk analysis using existing boosting libraries. To fill this gap, this paper introduces a new zero-inflated Poisson boosted tree model, which considers the relationship between the inflation probability $p$ and the distribution mean $\mu$, to be used with insurance claim data, including telematics data, that has a zero-inflated nature, in order to create a frequency model. We construct two zero-inflated models: one where the inflation probability is a function of the distribution mean and the other with no correlation. These models will be compared with other models, such as Poisson boosted tree, zero-inflated Poisson and Poisson GLM.

Boosting is a fitting process that iteratively combines weak learners into a strong learner to improve the accuracy of predictions. The first practical boosting algorithm, AdaBoost.M1, was developed by \citet{freund1997decision} for building precise binary classifiers. To handle regression problems, \citet{friedman2001greedy} introduced gradient boosting. Among the gradient boosting libraries that exist today, \texttt{XGBoost}, \texttt{LightGBM}, and \texttt{CatBoost} are the most popular. \texttt{XGBoost}, originally a research project by \citeauthor{chen2016xgboost} in 2014, has been recognized due to its successful application of the Gradient Boosting Decision Tree (GBDT) technique. Microsoft, in response to \texttt{XGBoost}'s success, introduced \texttt{LightGBM} (\citet{ke2017lightgbm}), an enhancement over \texttt{XGBoost}'s features to speed up the implementation of the GBDT method. Additionally, \texttt{CatBoost}, a library recently developed by Yandex (\citet{prokhorenkova2018catboost}), has also risen to prominence due to its ability to handle heterogeneous data, such as insurance data. In our research, we aim to contrast these libraries to identify the most suitable one for insurance claim data and to fit suggested actuarial frequency models to the data. While there have been attempts to compare the performance of these three libraries in implementing gradient boosting algorithms in other fields (\citet{al2019comparison}; \citet{bentejac2021comparative}), related studies are limited in actuarial science.

The structure of the remainder of this paper is as follows: Section 2 reviews the generic gradient boosting algorithm and provides an overview of \texttt{XGBoost}, \texttt{LightGBM}, and \texttt{CatBoost}. In Section 3, we present the zero-inflated Poisson boosted tree models, which assume a connection between the parameters. Section 4 is devoted to the application of these models to two auto insurance claim datasets. Finally, Section 5 offers brief concluding remarks on the study.

\section{Gradient Boosting Machine} \label{sec:GBM}

Boosting machines are believed to be among the most powerful learning algorithms discovered in recent years, quickly gaining popularity among actuaries. Boosting is an iterative fitting procedure that combines weak learners - those slightly better than random - into a strong learner for improved and more accurate predictions. Boosting can also be described as a stage-wise additive model, as one new weak learner is added at a time, and existing weak learners in the model are fixed and left unchanged. The first practical boosting algorithm, referred to as \texttt{AdaBoost.M1}, was introduced by \citet{freund1997decision}. Initially, boosting was intended for classification problems; however, the idea has since been expanded to include regression problems (\citet{hastie2009}). Gradient boosting, a boosting-like algorithm for regression, was introduced by \citet{friedman2001greedy}. Gradient boosting constructs a strong learner through a numerical optimization, where the objective is to minimize the model's loss by adding weak learners using a gradient descent procedure. 

\subsection{Generic algorithm} \label{sub:generic}

Given a training dataset $D = \{ \bm{x}_i, y_i\}^N_1$, gradient boosting iteratively constructs a sequence of functions of input variables, $F_0, F_1,\cdots, F_T$ , by minimizing the expected value of a given loss function, $L(y_i,F_t)$. Here, the loss function has two input values, the i-th output value $y_i$, and the t-th function $F_t$ that estimates $y_i$.
Assuming we have constructed function $F_t$ we can improve our estimates of $y_i$ by finding another function $F_{t+1} = F_t + f_{t+1}$ such that $f_{t+1}$ minimizes the expected value of the loss function. That is,

\begin{equation} \label{eq:1}
f_{t+1}=\argmin_{f \in H} \mathbb{E} L(y, F_{t+1}),
\end{equation}
where $H$ represents the set of candidate decision trees being evaluated, with the goal being to select one to add to the model. Furthermore, given the definition of $F_{t+1}$, it is possible to express the expected value of the loss function $L$ in terms of $F_t$ and $f_{t+1}$:

\begin{equation}  \label{eq:2}
\mathbb{E} L(y, F_{t+1})=\mathbb{E} L(y, F_t+f_{t+1}).
\end{equation}

According to Equation (\ref{eq:2}), we wish to minimize the expected value of the loss function, $L$, with respect to $y$ and $F_t$, while also considering an additional factor, $f_{t+1}$. Assuming $L$ is continuous and differentiable, the strategy is to adjust $F_t$ in the direction that $L$ decreases the most, which is associated with the rate of change of $L$. Thus, if $f_{t+1}$ is set in the direction where the gradient of $L$ with respect to $F_t$ is decreasing fastest, it results in the $f_{t+1}$ value that approximates the minimum of $\sum_{i=1}^{N}  L(y_i, F_{t+1}(\bm{x}_i))$. Under these assumptions then, we can write a reasonable approximation for $f_{t+1}$:

\begin{equation} \label{eq:3}
	f_{t+1} \approx \argmin_{f \in H} \mathbb{E} \left( \frac{\partial L}{\partial F_t}-f\right)^2. 
\end{equation}

Therefore, each $f$ can be seen as a greedy step in a gradient descent optimization for $F$. For that, each model, $f_{t+1}$, is trained on a new dataset $D = \{ \bm{x}_i, g_{i,t+1}\}^N_1$ where working response, $g_{i,t+1}$, are calculated by

\begin{equation} \label{eq:4}
\displaystyle g_{i,t+1}=\partial_{F_{t}} L(y_i,F_{t}(\bm{x}_i))=\frac{\partial L(y_i,F_t(\bm{x}_i))}{\partial F_t(\bm{x}_i)}.
\end{equation}

Among the gradient boosting libraries introduced to date, the most popular ones are \texttt{XGBoost}, \texttt{LightGBM}, and \texttt{CatBoost}. They make refinements to the generic algorithm described above. Researchers with a comprehensive understanding of how each library implements the GBDT technique will be better equipped to apply it across various fields. Therefore, we will provide detailed overviews of these libraries, along with a comparative analysis, in the following section.

\subsection{Comparison between \texttt{XGBoost}, \texttt{LightGBM} and \texttt{CatBoost}} \label{sub:comparison}

\texttt{XGBoost}(eXtreme Gradient Boosting) was originally developed as a research project by \citeauthor{chen2016xgboost} in 2014. Its successful implementation of the GBDT technique led to widespread recognition. In response to \texttt{XGBoost}'s success, Microsoft introduced \texttt{LightGBM}(Light Gradient Boosting Machine, \citet{ke2017lightgbm}), a tool designed to enhance certain features of \texttt{XGBoost} in order to speed up the GBDT implementation. In addition, \texttt{CatBoost}(Categorical Boosting), a library developed by Yandex (\citet{prokhorenkova2018catboost}), has gained recognition. As the name suggests, it offers a gradient boosting framework that excels in learning problems with heterogeneous features, adeptly managing categorical features. Through a detailed examination of \texttt{XGBoost}, \texttt{LightGBM}, and \texttt{CatBoost}, we will uncover the commonalities and differences between \texttt{XGBoost}, \texttt{LightGBM}, and \texttt{CatBoost}.

\textbf{Commonality.} These libraries focus solely on decision trees as weak learners. They provide an array of built-in loss functions, including, but not limited to, Poisson and Tweedie loss. For researchers wishing to use their own loss functions, these libraries facilitate the use of any user-defined loss function that can be implemented by defining a function that outputs both the gradient and the hessian(second-order gradient). To train the set of decision trees in the model while avoiding overfitting, a regularized loss function is employed to control the complexity of the trees:

\begin{equation} \label{eq:5}
	L=\sum_{i=1}^{N} L(y_i, F_T(\bm{x}_i)) +\sum_{j=1}^{T} w(f_j),
\end{equation}
where $w(f_j)$ is the regularization term, penalizing the complexity of the tree $f_j$. At the t-th iteration, a regularized loss function is defined as:

\begin{equation} \label{eq:6}
	L^t=\sum_{i=1}^{N} L(y_i, F_t(\bm{x}_i)) +\sum_{j=1}^{t} w(f_j),
\end{equation}
with

\begin{equation} \label{eq:7}
	F_t(\bm{x}_i)=\sum_{j=1}^{t}f_j(\bm{x}_i) =F_{t-1}(\bm{x}_i)+f_t(\bm{x}_i).
\end{equation}

As explained in section \ref{sub:generic}, we train one decision tree and add it to the existing model at each iteration. Therefore, equation (\ref{eq:6}) can be written as:

\begin{equation} \label{eq:8}
	L^t=\sum_{i=1}^{N} L(y_i,F_{t-1}(\bm{x}_i)+f_t(\bm{x}_i)) + w(f_t) +\text{constant}.
\end{equation}

Taking the Taylor expansion of the loss function up to the second order and removing all the constants, the t-th regularized loss function can be simplified as:

\begin{equation} \label{eq:9}
	L^t=\sum_{i=1}^{N} \left[ g_{i,t} f_t(\bm{x}_i) + \frac{1}{2} h_{i,t} f_t(\bm{x}_i)^2\right] +w(f_t),
\end{equation}
where $g_{i,t}=\partial_{F_{t-1}} L(y_i,F_{t-1}(\bm{x}_i))$ and $h_{i,t}=\partial^2_{F_{t-1}} L(y_i,F_{t-1}(\bm{x}_i))$.
One significant advantage of this definition is that the value of the loss function only depends on $g_{i,t}$ and $h_{i,t}$. This is the mechanism through which these libraries support user-defined loss functions.

\textbf{Difference.} These libraries are differentiated based on three aspects: the methods of tree splitting, the handling of categorical features, and the tree growth strategies. 

	\setlength\itemsep{0.00001em}
\begin{itemize}
	\setlength\itemsep{0.001em}
	\item Methods of tree splitting: Prior to training, all libraries generate feature-split pairs for all features, examples of which include (credit score $<$700), or (insured age $<$ 30). \texttt{XGBoost} uses a Histogram-based algorithm to compute the best split, \texttt{LightGBM} employs Gradient-based One-Side Sampling(GOSS), and \texttt{CatBoost} introduces a technique known as Minimal Variance Sampling(MVS). 
	
	The Histogram-based algorithm splits all values for a feature into discrete bins and uses these bins to identify the split value. In terms of training speed, while it is more efficient than enumerating all possible split points on the pre-sorted feature values, it lags behind both GOSS and MVS.
	
	GOSS is a sampling method which downsamples the observations based on their gradients.  Observations with smaller gradients are typically well-trained, and those with larger gradients are undertrained. A straightforward downsampling approach would be to remove observations with small gradients, concentrating exclusively on observations with large gradients; however, this approach could skew the data distribution. As a result, GOSS retains observations with larger gradients while performing random sampling on observations with smaller gradients.
	
	MVS, on the other hand, conducts weighted sampling at the tree level, as opposed to the split level. MVS samples observations such that those with the largest gradient values are selected with certainty, while each other instance is sampled with a probability proportional to its gradient. MVS can be viewed as an enhanced version of GOSS (for more details, see \citet{johnson2018training}).
	
	\item Handling of categorical features: When dealing with categorical features during splitting, it's important to note that these categories don't inherently possess an order. Consequently, we need different strategies. Ideally, we aim to group categories within a categorical feature that generate similar leaf values.
	
	\texttt{LightGBM} and \texttt{XGBoost} use a technique known as optimal partitioning to split categories. The primary concept is to sort the categories according to the loss function at each split and then enumerate these sorted values to identify the best split.
	
	Contrastingly, \texttt{CatBoost} adopts a method referred to as ``Ordered Target Statistic" for encoding categorical features. This characteristic distinguishes \texttt{CatBoost} from other libraries. As a result, it has proven to be the most effective library for GBDT implementation when dealing with heterogeneous datasets, which contain a mix of numerical and categorical features, such as auto telematics datasets. 
	A target statistic, in this context, is a value computed from the response values associated with a specific category of a categorical feature in a dataset. It is typically the expected target $y$, conditioned by the category. An efficient and effective method to process a categorical feature $k$ is to substitute the category $x^k_i$ of the $i$-th training observation with the target statistic:
	\begin{equation} \label{eq:10}
		\hat{x}^k_i=\frac{\sum_{\bm{x}j \in D_k} [\mathds{1}{{x^k_j=x^k_i}}]y_j+ap}{\sum_{\bm{x}j \in D_k} [\mathds{1}{{x^k_j=x^k_i}}]+a}.
	\end{equation}
	Here, $a > 0$ is a parameter, and $D_k$ is the subset of the whole dataset excluding $\bm{x}_i$. If we exclude terms involving $p$ and $a$, it merely computes the conditional expectation of the target $y$. The term $p$ is included for smoothing, and to prevent target leakage, we compute the target statistic solely using observations from $D_k$ (for more details, refer to \citet{prokhorenkova2018catboost}).
	
	\begin{figure}[htbp]
	\centering
	\includegraphics[width=\textwidth]{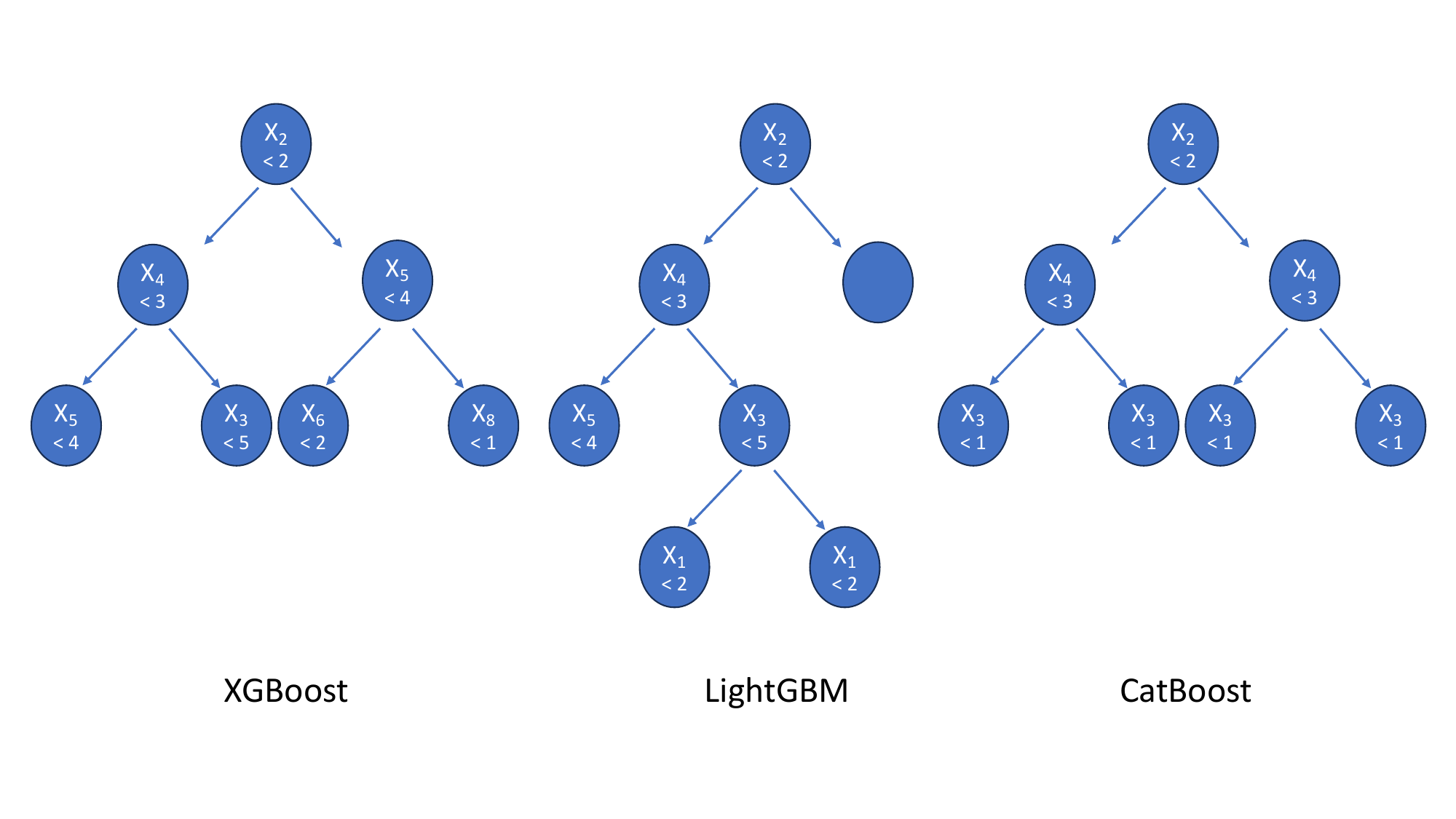}
	\caption{Illustrations of different tree growth: Level-wise tree growth (left), Leaf-wise tree growth (middle), and Oblivious tree growth (right)}
	\label{fig1:tree_growth}
\end{figure}

	\item Tree growth strategies: \texttt{XGBoost} grows trees up to a specified ``max\_depth" hyperparameter, after which it starts pruning the tree backwards, removing splits that do not contribute to a positive gain. In \texttt{XGBoost}, trees grow level-wise.
	
	In contrast, \texttt{LightGBM} employs a leaf-wise tree growth strategy. Rather than examining all previous leaves for each new leaf, as depicted in Figure \ref{fig1:tree_growth}, it chooses to grow the leaf that minimizes the loss, thus allowing for the growth of an imbalanced tree. Because it does not grow level-wise but leaf-wise, overfitting can occur when dealing with small data sets.
	
	\texttt{CatBoost}, on the other hand, uses oblivious trees (\citet{kohavi1994bottom}). These trees apply the same splitting criterion across an entire level of the tree, as shown in Figure \ref{fig1:tree_growth}. For each level of such a tree, the feature-split pair that contributes to the lowest loss is chosen and applied to all the level’s nodes. Such trees have balanced structures, are less susceptible to overfitting, and significantly speed up predictions during testing time.	
\end{itemize}

\section{Zero-Inflated Poisson Boosted Tree} \label{sec:sc2}

The Poisson distribution is a classical model used to represent claim frequency in the field of actuarial science. Assuming that the claim count is a random variable, denoted by $y$, and follows a Poisson distribution, its probability mass function can be expressed as:

\begin{equation} \label{eq:11}
	P(y|\mu)=\frac{\mu^y e^{-\mu}}{y!}, \;\; y=0,1,2,\cdots,
\end{equation}
where $\mu>0$ represents the expected value of $y$.

The boosted tree model assumes that the logarithm of the expected value of $y$, given $\bm{x}$, can be modeled by the prediction score from gradient boosting as follows:

\begin{equation} \label{eq:12}
	\ln \mathbb{E} (y\mid \bm {x} ) = \ln w + F_T(\bm{x}) ,
\end{equation}
\[	F_T(\bm{x})=f_1(\bm{x})+f_2(\bm{x})+\cdots +f_t(\bm{x})+\cdots+f_T(\bm{x}),\]
where $\ln w$ is the offset term and $f_t(\bm{x})$ is the $t$-th tree in the gradient boosting model. The Poisson boosted tree assumes that the response variable $y$ follows a Poisson distribution. Thus, $\mu$ in the Poisson distribution is equivalent to $ \mathbb{E} (y\mid \bm {x} )$ in equation (\ref{eq:12}). For traditional GLMs, the prediction score, $F_T(\bm{x}) $, corresponds to the linear combination of features.

Libraries such as \texttt{XGBoost}, \texttt{LightGBM}, and \texttt{CatBoost} have a built-in parameter that trains a Poisson boosted tree by minimizing the negative log-likelihood of the Poisson distribution as a loss function. However, they do not support models with an offset, necessitating a user-defined loss function if an offset term is required to reflect the concept of exposure in insurance.

Although we can readily train a Poisson boosted tree using the built-in parameter, this traditional distribution faces challenges when dealing with insurance claim data, particularly when overdispersion occurs due to the number of observed zeros exceeding the number expected under the distributional assumption. To handle datasets with excess zeros, researchers have employed ``zero-inflated" models. The zero-inflated Poisson (ZIP) distribution combines a point mass at zero and a Poisson distribution, improving the ability to estimate $\mu$ using information from zero-claim observations. If $y$ follows a ZIP distribution, the probability mass functions are defined as follows:

\begin{equation} \label{eq:13}
	P(y|\mu, p)=\begin{cases}
		p+(1-p)e^{-\mu} &\quad y=0\\
		(1-p)\displaystyle\frac{\displaystyle\mu^y e^{-\mu}}{y!} &\quad y=1,2,\cdots,\\ 
	\end{cases}
\end{equation}
where $p$ is inflation probability, the degree of inflation. The ZIP distribution has an expected value for $y$ of $(1 -p)\mu$, and a variance of $\mu(1 -p)×(1+p\mu)$, giving it over-dispersed characteristics.

\noindent In ZIP boosted tree, we define: 
\begin{equation} \label{eq:14}
	\ln \mu=\ln w + \displaystyle{F^{po}_T(\bm{x}) },
\end{equation}
\begin{equation} \label{eq:15}
\text{logit}(p)	=\ln\frac{p}{1-p} =  F^{logit}_T(\bm{x}). 
\end{equation}
As suggested by \citet{lambert1992zero}, the features that can affect the Poisson mean $\mu$ may be different from those that influence the inflation probability $p$. Even if the influencing features are the same, if $\mu$ and $p$ are independent, a ZIP boosted tree would require twice as many trees as a Poisson boosted tree and the final model would be the combination of two separate models, making it difficult to analyze the interaction effects between features and the effect of each feature on the claim prediction. This observation suggests reducing the number of trees and training one model by treating $p$ as a function of $\mu$. In the following sections, we will explore the implementation of the ZIP boosted tree, looking into the relation of two parameters.

\subsection{$p$ is a function of $\mu$} \label{sub:related}
We propose an assumption where $p$ is a function of $\mu$. Given minimal prior information about the relationship between $p$ and $\mu$, we suggest a refinement of the parameterization of $p$ and $\mu$, originally introduced by \citet{lambert1992zero} for GLM: 
\[	\ln \mu = \ln w + F_T(\bm{x}), \]
\[	\text{logit}(p)	=\ln\frac{p}{1-p} =  -\gamma F_T(\bm{x}). \]
From this, we can deduce that
\begin{equation} \label{eq:16}
p=\frac{1}{1+\mu^{\gamma}}.
\end{equation}
According to equation (\ref{eq:16}), for $\gamma >0$, the inflation probability decreases as $\mu$ increases. Conversely, when $\gamma <0$, $\mu$ increases as excess zeros become more likely, which is inconsistent with claim data. Consequently, $\gamma$ is predefined as a pre-training hyper-parameter with a positive value.

In the training of the ZIP boosted tree, we use the negative log-likelihood of the ZIP distribution as our loss function, which is expressed as:
 \begin{equation} \label{eq:17}
 L (y, F_T(\bm{x})) =\begin{cases}
 		-\ln(1+\mu^{\gamma}e^{-\mu})+\ln(1+\mu^{\gamma}) &\quad y=0\\
 -\gamma \ln \mu +\ln(1+\mu^{\gamma})+\mu-y\ln \mu  &\quad y=1,2,\cdots,\\ 
 \end{cases}
 \end{equation}
where $\mu=we^{F_T}$. The gradient and hessian of the loss function, as defined in Equation (\ref{eq:17}), with respect to $F_t$, are:

 \begin{equation} \label{eq:18}
g_{t+1}=\partial_{F_{t}} L(y,F_{t})= \begin{cases}
		\displaystyle \frac{\mu^{\gamma}e^{-\mu}(\mu-\gamma)}{1+\mu^{\gamma}e^{-\mu}}+\frac{\gamma \mu^{\gamma}}{1+\mu^{\gamma}}&\quad y=0\\
	     \displaystyle\frac{\gamma \mu^{\gamma}}{1+\mu^{\gamma}}+\mu-\gamma-y&\quad y=1,2,\cdots,\\ 
	\end{cases}
\end{equation}

 \begin{equation} \label{eq:19}
h_{t+1}=\partial^2_{F_{t}} L(y,F_{t})=\begin{cases}
	\displaystyle\frac{\mu^{\gamma}e^{-\mu}(\mu-(\gamma-\mu)^2)(1+\mu^{\gamma}e^{-\mu})+\mu^{2\gamma}e^{-2\mu}(\gamma-\mu)^2}{(1+\mu^{\gamma}e^{-\mu})^2} +\displaystyle\frac{\gamma^2\mu^{\gamma}}{(1+\mu^{\gamma})^2}&\quad y=0\\
	\displaystyle\frac{\gamma^2\mu^{\gamma}}{(1+\mu^{\gamma})^2}+\mu&\quad y=1,2,\cdots.\\ 
		\end{cases}
\end{equation}
For a detailed description of the training algorithm, please refer to Algorithm \ref{alg:case1} in  Appendix A.

\subsection{$p$ and $\mu$ are unrelated} \label{sub:unrelated}
Assuming that the influencing features are the same and that $\mu$ and $p$ are not functionally related, as suggested by equations (\ref{eq:14}) and (\ref{eq:15}), we find it necessary to train $F^{po}_T(\bm{x})$ and $F^{logit}_T(\bm{x})$ independently. Here, $F^{po}_T(\bm{x})$ indicates the prediction score for the mean parameter $\mu$, which is converted by an exponential transformation, while $F^{logit}_T(\bm{x})$ corresponds to the prediction score for the inflation probability $p$, transformed by a sigmoid function. The loss function is thus defined as:
 \begin{equation} \label{eq:20}
	L(F^{po}_T(\bm{x}), F^{logit}_T(\bm{x})) =
		\begin{cases}
	-\ln \left(p+(1-p)e^{-\mu}\right) &\quad y=0\\
-\ln(1-p) - y\ln\mu  +\mu &\quad y=1,2,\cdots,\\ 
	\end{cases}
\end{equation}
where $\mu=we^{F^{po}_T}$ and $p=\text{logit}^{-1}(F^{logit}_T(\bm{x}))$. The gradients and hessians of the loss function, as defined in Equation (\ref{eq:20}), with respect to $F^{po}_t$ are:

\begin{equation} \label{eq:21}
	g^{po}_{t+1}=\partial_{F^{po}_{t}} L(y,F^{po}_{t},F^{logit}_{t})= \begin{cases}
	\displaystyle\frac{(1-p)\mu e^{-\mu}}{p+(1-p)e^{\mu}}	&\quad y=0\\
	\mu-y	&\quad y=1,2,\cdots,\\ 
	\end{cases}
\end{equation}

\begin{equation} \label{eq:22}
	h^{po}_{t+1}=\partial^2_{F^{po}_{t}} L(y,F^{po}_{t},F^{logit}_{t})=\begin{cases}
		\displaystyle\frac{(1-p)\mu e^{-\mu}(1-\mu)}{p+(1-p)e^{\mu}}+	\displaystyle\left(\frac{(1-p)\mu e^{-\mu}}{p+(1-p)e^{\mu}}\right)^2&\quad y=0\\
		\mu&\quad y=1,2,\cdots.\\ 
	\end{cases}
\end{equation}
The gradients and the hessians of the loss function, as defined in Equation (\ref{eq:20}), with respect to $F^{logit}_t$ are defined as: 
\begin{equation} \label{eq:23}
	g^{logit}_{t+1}=\partial_{F^{logit}_{t}} L(y,F^{po}_{t+1},F^{logit}_{t})= \begin{cases}
		\displaystyle\frac{1}{1+e^{\mu+F^{logit}_{t}}}-\frac{1}{1+e^{F^{logit}_{t}}}&\quad y=0\\
		\displaystyle\frac{e^{F^{logit}_{t}}}{1+e^{F^{logit}_{t}}}&\quad y=1,2,\cdots,\\ 
	\end{cases}
\end{equation}

\begin{equation} \label{eq:24}
	h^{logit}_{t+1}=\partial^2_{F^{logit}_{t}} L(y,F^{po}_{t+1},F^{logit}_{t})=\begin{cases}
		\displaystyle\frac{e^{F^{logit}_{t}}}{\left(1+e^{F^{logit}_{t}}\right)^2}-\frac{e^{\mu+F^{logit}_{t}}}{\left(1+e^{\mu+F^{logit}_{t}}\right)^2}&\quad y=0\\
		\displaystyle\frac{e^{F^{logit}_{t}}}{\left(1+e^{F^{logit}_{t}}\right)^2}&\quad y=1,2,\cdots.\\ 
	\end{cases}
\end{equation}
where $\mu=we^{F^{po}_{t+1}}$.

To minimize the loss function using two separate models, we adopt the methodology proposed by \citet{meng2022actuarial}. This method uses the principle of coordinate descent optimization to enhance the predictive model for a specific parameter, keeping the score of the other parameter fixed. By following alternating cycles, only one direction corresponding to the estimated parameter is used for the Taylor expansion, allowing for the update of the boosted trees as delineated in Section \ref{sub:generic}. For a detailed description of the training algorithm, please refer to Algorithm \ref{alg:case2}, Appendix A.

\section{Experimental results } \label{sec:empirical}
This section evaluates the model proposed in section \ref{sec:sc2} and presents a comprehensive comparative analysis of \texttt{XGBoost}, \texttt{LightGBM}, and \texttt{CatBoost} models. The analysis aims to identify which claim frequency models are best suited for handling excess zero auto claim datasets. The analyses are based on two datasets: the French Motor Third-Party Liability (MTPL) dataset, accessible through the R library CASdatasets, and a synthetic telematics dataset provided by \citet{so2021synthetic}. The French Motor Third-Party Liability (MTPL) dataset, consisting of 678,013 policies, exhibits a zero-inflation trait as only 34,060 policies incurred at least one claim. Similarly, the synthetic telematics dataset, which comprises 100,000 policies, shows this zero-inflation trait, with only 2,698 policies incurring at least one claim.

In total, eleven models were trained for this study. Three models per library were investigated, including Poisson boosted tree (PB), ZIP boosted tree with only $\mu$ modeled and $p$ calculated as a function of $\mu$ (ZIPB1), and ZIP boosted tree with both $\mu$ and $p$ modeled (ZIPB2). Poisson GLM (PG) and zero-inflated Poisson GLM (ZIPG) were implemented using the ``statsmodels" Python package. While the built-in loss function was used for most models, ZIPB1 and ZIPB2 were implemented with a custom loss function to apply the training methods outlined in Algorithms 1 and 2 as presented in the Appendix A.

The hyperparameters for the boosted trees were set as follows: we determined the number of trained trees (T) to be 500, chose the learning rate from the grid values $\alpha=\{0.01,0.05,0.1\}$, and selected the L2 penalty parameter from the grid values $\lambda=\{0,100,\ldots,500\}$. For the ZIPB1 model, an additional hyperparameter, $\gamma$, was introduced. It was likewise determined using grid search from the values $\gamma=\{1,5,10,50,100,500\}$. 
In addition, the maximum number of trees was constrained to 8 for \texttt{XGBoost} and \texttt{CatBoost}. \texttt{LightGBM}, in contrast, uses a leaf-wise tree growth strategy. Therefore, we restricted the maximum number of leaves to $2^8=256$ instead of limiting the number of trees. For categorical features, models were trained using each library's unique treatment approach, as described in Section \ref{sub:comparison}, with the exception of binary categorical features, for which we used one-hot encoding.

The model comparisons were performed based on prediction accuracy, which in our context refers to the model's ability to distinguish between policyholders with varying risk levels. The metrics used for measuring prediction accuracy will be described in the following subsection. For these analyses, 20\% of the total dataset was reserved as the test set. The optimal hyperparameters were determined using a 3-fold cross-validation method applied to the remaining training dataset.

\subsection{Performance evaluation} \label{sub:metric}
The comparison between Poisson and zero-inflated Poisson models can pose challenges due to their unique assumptions and differing structures. To effectively assess and contrast these models, the paper uses a variety of evaluation metrics. These metrics comprise Deviance / Pseudo R-squared, the Vuong test, and Randomized Quantile Residuals (RQR).

\subsubsection{Deviance / Pseudo R-squared }\label{subsub:metric1}
In statistical modeling, Deviance measures the discrepancy in log-likelihood between the evaluated model and the saturated model - a model that perfectly generates the observations, such as $\hat{\mu}_i=y_i$.

For zero-inflated Poisson models, the unit deviance for the i-th observation can be computed using the following equation:
\[d(y_i,(\hat{\mu}_i, \hat{p}_i))=\begin{cases} 
-2	 \ln\left(\hat{p}_i + (1 - \hat{p}_i) \exp(-\hat{\mu}_i) \right) &  y_i = 0 \\
2\left( 	y_i \ln y_i  - y_i - \ln\left(1 - \hat{p}_i \right) -y_i \ln \hat{\mu}_i  +\hat{\mu}_i  \right)& y_i=1,2,\cdots.
\end{cases}\]

The Mean Deviance $D(y, (\hat{\mu}, \hat{p}))$ for a given model is the average of the unit deviances across all observations. Using this metric, we can compute McFadden’s Pseudo R-squared, a measure that captures the degree of model improvement over the null model (a model without predictors), as follows:
\[\text{pseudo } R^2=1-\frac{D(y, (\hat{\mu}, \hat{p}))}{D(y, (\bar{y}, \bar{p}))}. \]

For this calculation, we assume that $\bar{p}=0.5$. A model is generally preferred if it yields lower deviance and higher Pseudo R-squared values. For the Poisson model, these metrics are computed by setting both $\hat{p}$ and $\bar{p}$ to zero.

\subsubsection{Vuong test}\label{subsub:metric2}
The Vuong test, introduced by \citet{vuong1989likelihood}, provides a robust method for comparing the zero-inflated Poisson model with other non-nested models for count data, such as the Poisson model. Let $P_K(y_i|\bm{x}_i)$ represent the probability distribution functions for the i-th observation from model $K$. If both models fit the data with equal efficacy, their respective likelihood functions would be almost identical. In contrast, any differences between the likelihood functions can indicate which model provides a better fit. The Vuong test is grounded in this concept.
We define $m_i$ as follows:
\[m_i=\ln \displaystyle\frac{P_1(y_i|\bm{x}_i)}{P_2(y_i|\bm{x}_i)}.\]
The Vuong statistic for the null hypothesis $\mathbb{E} (m_i)=0$ is given by:
\[V=\displaystyle\frac{\sqrt{n}\left(\displaystyle\frac{1}{n}\sum_{i=1}^{n}m_i \right) }{\sqrt{\displaystyle\frac{1}{n}\sum_{i=1}^{n}(m_i-\bar{m})^2}}.\]
Under the null hypothesis, the Vuong statistic follows an asymptotic normal distribution. At a 5\% significance level, if $V > 1.96$, the first model is preferred; if $V < -1.96$, the second model is preferred.

\subsubsection{Randomized Quantile Residuals}\label{subsub:metric3}
Model goodness-of-fit is often visually assessed using both Pearson and deviance residuals, which are approximately standard normally distributed when the model fits the data adequately. However, as pointed out by \citet{feng2020comparison}, these residuals diverge from normality when dealing with discrete response variables. To address the limitations of traditional residuals, \citet{dunn1996randomized} introduced the method of Randomized Quantile Residuals (RQRs). \citet{feng2020comparison} examined the normality of RQRs and compared its performance with traditional residuals in diagnosing count regression models. Through a series of simulation studies, they demonstrated that RQRs can more advantageously detect various forms of model misspecification for count regression models than traditional residuals. Thus, to ascertain the most suitable model for auto claim data in our paper, we will use RQRs.

The RQRs for the Poisson and zero-inflated Poisson models can be formulated as follows:

\noindent For the Poisson model:
\[r_i^{po}= \Phi^{-1}(F_{y|\hat{\mu}}(y_i-1) + u_i \cdot f_{y|\hat{\mu}}(y_i)).\]

\noindent For the zero-inflated Poisson model:
\[r_i^{zip}= \begin{cases} 
	 \Phi^{-1}( (\hat{p}_i + (1-\hat{p}_i) \cdot e^{-\hat{\mu}_i} ) \cdot u_i )&  y_i = 0 \\
    \Phi^{-1}((\hat{p}_i + (1-\hat{p}_i) \cdot e^{-\hat{\mu}_i} ) + (1-\hat{p}_i ) \cdot F_{y|\hat{\mu}}(y_i-1) + u_i\cdot (1-\hat{p}_i ) \cdot f_{y|\hat{\mu}}(y_i))& y_i=1,2,\cdots.
\end{cases}\]
In these formulations, $u_i$ is a random number drawn from a uniform distribution on the interval (0,1), while $F_{y|\hat{\mu}}$ and $f_{y|\hat{\mu}}$ represent the cumulative distribution function (CDF) and the probability mass function (PMF) of the Poisson distribution, respectively.

\subsection{French Motor Third-Party Liability} \label{sub:data1}

The French Motor Third-Party Liability (MTPL) dataset is a publicly available auto insurance portfolio used for claim frequency modeling. This dataset, named \texttt{freMTPL2freq}, can be accessed through the R library CASdatasets as referenced in \citet{vignette2016reference}. It comprises a French Motor Third-Party Liability (MTPL) insurance portfolio, recording claim counts observed over one accounting year. The dataset includes a total of 678,013 policies, of which only 34,060 incurred at least one claim, thus exhibiting a characteristic zero-inflation. We fit the claim frequency models using exposure and nine features, which include information about the driver and vehicle. Among these nine features, four are categorical, each consisting of 6, 11, 2, and 22 categories respectively. This dataset, therefore, features high-cardinality categorical variables, pointing out the need for suitable model selection capable of effectively handling such features for optimal performance and accurate data fitting.  A detailed description of the dataset is provided in Table \ref{tab:VD1}, Appendix B.

\begin{table}[htbp]
	\caption{Comparison of Pseudo $\text{R}^2$ and Deviance for French Motor Third-Party Liability Data Across 11 models}
	\centering
	\resizebox{!}{1.1cm}{
		\begin{tabular}{l|c|c|c|c|c|c|c|c|c|c}
			\hline
			& \multicolumn{5}{c|}{Pseudo $\text{R}^2$} & \multicolumn{5}{c}{Deviance} \\ \cline{2-11}
			& ZIPB2 & ZIPB1 & PB & ZIPG  &PG & ZIPB2 & ZIPB1 &PB & ZIPG&PG  \\ \hline
			\texttt{XGBoost} &0.455&0.273&0.136&\multirow{3}{*}{0.046}&&1.100&1.467&1.908&\multirow{3}{*}{1.925}&\\ \cline{1-4}\cline{7-9}
			\texttt{LightGBM}&0.453&0.291&0.141&\multirow{3}{*}{}&0.026&1.105&1.431&1.897&\multirow{3}{*}{}&2.151\\\cline{1-4}\cline{7-9}
			\texttt{CatBoost} &\color{red}0.520&0.292&0.140&&&\color{red}0.970&1.430&1.898&&\\ \hline
	\end{tabular}} \label{tab:metric freMTPL}
\end{table}

We trained eleven models on the data. The performance of each model, as indicated by Pseudo $R^2$ and mean Deviance, is presented in Table \ref{tab:metric freMTPL}. Among all, the \texttt{CatBoost} ZIPB2 model excelled with the highest Pseudo $R^2$ and the lowest mean Deviance, underlining its superior performance. All ZIPB2 models outperformed ZIPB1 models in terms of Pseudo $R^2$ and Deviance, with the \texttt{CatBoost} model showing the best performance. Additionally, all ZIPB1 models had better metrics than PB, ZIPG, and PG models. The PG model had the worst performance, with the lowest Pseudo $R^2$ and highest Deviance.

The Vuong statistics for pairs of the 11 models, with corresponding p-values, are represented in Table \ref{tab:vuong}. A significance level for the p-value was set at 0.05. A first model is considered significantly superior to a second model if the Vuong statistic is positively large with a p-value less than 0.05. Based on Table \ref{tab:vuong}, the ranking of models' performance is as follows: \texttt{CatBoost} ZIPB2$>$\texttt{LightGBM} ZIPB2, \texttt{XGBoost} ZIPB2$>$\texttt{CatBoost} ZIPB1, \texttt{LightGBM} ZIPB1, \texttt{XGBoost} ZIPB1$>$\texttt{CatBoost} PB, \texttt{LightGBM} PB, \texttt{XGBoost} PB, ZIPG$>$PG.

\begin{table}[htbp]
	\caption{Vuong Statistics and Corresponding P-values for French Motor Third-Party Liability Data}\label{tab:vuong}
	\centering
	\begin{threeparttable}
		\resizebox{\textwidth}{!}{
			\begin{tabular}{ccc|rrr|rrr|rrr|r|r}
				\multicolumn{14}{c}{Second Model} \\ 
				\toprule
				&&& \multicolumn{3}{c|}{\texttt{XGBoost}} & \multicolumn{3}{c|}{\texttt{LightGBM}}& \multicolumn{3}{c|}{\texttt{CatBoost}}  \\
				&&& ZIPB2 & ZIPB1 & PB & ZIPB2 & ZIPB1 & PB & ZIPB2 & ZIPB1 & PB & ZIPG & PG \\ \cline{2-14}
				\multirow{11}{*}{\rotatebox[origin=c]{90}{First Model}} &\multirow{3}{*}{\texttt{XGBoost}}&ZIPB2 &  & ... & ... & ... & ... & ... & ... & ... & ... & ... & ... \\
				&\multirow{3}{*}{}&ZIPB1                &-13.1(0)   && ... & ... & ... & ... & ... & ... & ... & ... & ... \\
				&&PB                                                 & -20.3(0) &-21.0(0)&  & ... & ... & ... & ... & ... & ... & ... & ... \\\cline{2-14}
				&\multirow{3}{*}{\texttt{LightGBM}}&ZIPB2 & -0.2(\color{red}0.8) &11.7(0) &17.8(0)&  & ... & ... & ... & ... & ... & ... & ... \\
				&\multirow{3}{*}{}&ZIPB1                 &-8.1(0)  &1.6(\color{red}0.1)&15.9(0)& -7.4(0) &  & ... & ... & ... & ... & ... & ... \\
				&&PB                                                 &-18.8(0)  &-18..8(0)&0.8(\color{red}0.4) &-17.8(0)&-13.6(0)&  & ... & ... & ... & ... & ... \\\cline{2-14}
				&\multirow{3}{*}{CatBoost}&ZIPB2 &10.1(0)  &13.7(0)&18.6(0)&6.5(0)&9.6(0.002)&17.7(0)&&...&...&...&...\\
				&\multirow{3}{*}{}	&ZIPB1              &-8.1(0) &1.6(\color{red}0.1)&15.9(0)&-7.4(0)&0.4(\color{red}0.7)&13.6(0)&-9.6(0)&  & ... & ... & ... \\
				&&PB                                          &-20.4(0)  &-20.5(0)&1.3(\color{red}0.2)&-18(0)&-16.1(0)&-0.1(\color{red}0.9)&-18.8(0)&-15.1(0)&& ... & ... \\\cline{2-14}
				&&ZIPG          &-13.1(0) & -11.0(0) &-0.5(\color{red}0.6)&-12.3(0)&-17.0(0)&-0.7(\color{red}0.5)&-13.2(0)&-16.9(0)&-0.7(\color{red}0.5)&& ... \\ \cline{3-14}
				&&PG                                                 &-18.2(0) &-18.5(0) &-7.6(0)&-17.0(0)&-27.6(0)&-7.6(0)&-17.5(0)&-27.7(0)&-8.2(0)&-31.6(0)&\\
				\bottomrule
		\end{tabular}} 
		\begin{tablenotes}
			\small
			\item Numbers represent Vuong statistics, with p-values shown in parentheses.
		\end{tablenotes}
	\end{threeparttable}
\end{table}

\begin{figure}[htbp]
	\centering
	\includegraphics[scale=0.35]{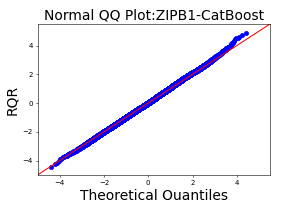}
	\includegraphics[scale=0.35]{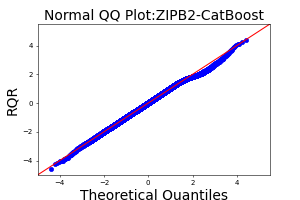}
	\includegraphics[scale=0.35]{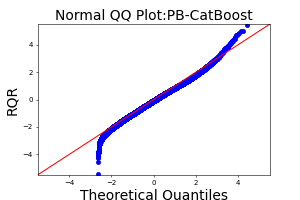}
	\includegraphics[scale=0.35]{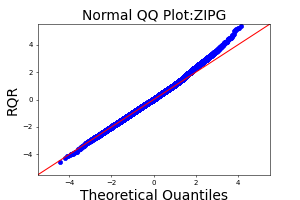}
	\includegraphics[scale=0.35]{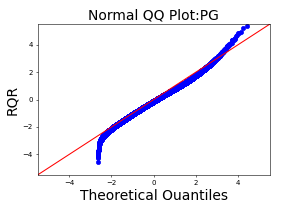}
	\caption{Q-Q Plots of Randomized Quantile Residuals: Five Models for French Motor Third-Party Liability Data - \texttt{CatBoost} ZIPB1 (upper left), \texttt{CatBoost} ZIPB2 (upper middle), \texttt{CatBoost} PB (upper right), ZIPG (bottom left), PG (bottom right)}	\label{fig7:RQR_fre}
\end{figure}

Figure \ref{fig7:RQR_fre} presents a Normal Q-Q plot of Randomized Quantile Residuals for five models: \texttt{CatBoost} ZIPB1, \texttt{CatBoost} ZIPB2, \texttt{\texttt{CatBoost}} PB, ZIPG, and PG, fitted to the French MTPL. The plot demonstrates significant improvements in the tail of the estimated residuals of ZIPB1, ZIPB2, and ZIPG when compared with PB and PG. Especially, for ZIPB2 and ZIPB1, almost all samples are aligned with the standard normal distribution. This observation emphasizes the superior fit of zero-inflated models over traditional counterparts for this specific dataset. Consequently, the zero-inflated Poisson boosted tree model with independent parameters yields the best results, followed by the zero-inflated Poisson boosted tree model with related parameters, and \texttt{\texttt{CatBoost}} outperforms other libraries.

\subsection{Synthetic telematics data} \label{sub:data2}
The synthetic telematics dataset, produced by \citet{so2021synthetic}, is publicly accessible at \url{http://www2.math.uconn.edu/~valdez/data.html}. As modeled on real data from a Canadian insurer, this dataset was generated using the Synthetic Minority Oversampling Technique (SMOTE) and a feedforward Neural Network (NN). It includes 100,000 data samples, of which only 2,698 policies incurred at least one claim, thereby demonstrating a zero-inflation trait. The dataset consists of 52 features categorized into Traditional data (such as insured age and gender), Telematic data (including total miles driven, harsh acceleration, harsh braking), and Response Data (claim counts and claim amounts). The claim frequency models were fit using exposure and 49 features.  Among these 49 features, five are categorical, comprising 2, 2, 4, 2, and 55 categories respectively. Thus, this dataset features a high-cardinality categorical attribute, emphasizing that the selection of models handling such features well is essential for optimal performance and accurate data fitting. A comprehensive description of this dataset is provided in Table \ref{tab:VD2}, Appendix B.
\begin{table}[htbp]
	\caption{Comparison of Pseudo $\text{R}^2$ and Deviance for Synthetic Telematics Data Across 11 Models}
	\centering
	\resizebox{!}{1.1cm}{
		\begin{tabular}{l|c|c|c|c|c|c|c|c|c|c}
			\hline
			& \multicolumn{5}{c|}{Pseudo $\text{R}^2$} & \multicolumn{5}{c}{Deviance} \\ \cline{2-11}
			& ZIPB2 & ZIPB1 & PB & ZIPG  &PG & ZIPB2 & ZIPB1 &PB & ZIPG&PG  \\ \hline
			\texttt{XGBoost} &0.364&0.311&0.291&\multirow{3}{*}{0.144}&&0.214&0.232&0.231&\multirow{3}{*}{0.288}&\\ \cline{1-4}\cline{7-9}
			\texttt{LightGBM}&0.386&0.356&0.329&\multirow{3}{*}{}&0.106&0.207&0.217&0.219&\multirow{3}{*}{}&0.291\\\cline{1-4}\cline{7-9}
			\texttt{CatBoost} &0.390&\color{red}0.419&0.341&&&0.205&\color{red}0.196&0.215&&\\ \hline
	\end{tabular}} \label{tab:metric STD}
\end{table}
\begin{table}[htbp]
	\caption{Vuong Statistics and Corresponding P-values for Synthetic Telematics Data}\label{tab:vuong STD}
	\centering
	\begin{threeparttable}
		\resizebox{\textwidth}{!}{
			\begin{tabular}{ccc|rrr|rrr|rrr|r|r}
				\multicolumn{14}{c}{Second Model} \\ 
				\toprule
				&&& \multicolumn{3}{c|}{\texttt{XGBoost}} & \multicolumn{3}{c|}{\texttt{LightGBM}}& \multicolumn{3}{c|}{\texttt{CatBoost}}  \\
				&&& ZIPB2 & ZIPB1 & PB & ZIPB2 & ZIPB1 & PB & ZIPB2 & ZIPB1 & PB & ZIPG & PG \\
				\cline{2-14}
				\multirow{11}{*}{\rotatebox[origin=c]{90}{First Model}} &\multirow{3}{*}{\texttt{XGBoost}}&ZIPB2 &  & ... & ... & ... & ... & ... & ... & ... & ... & ... & ... \\
				&\multirow{3}{*}{}&ZIPB1                &-7.0(0)   && ... & ... & ... & ... & ... & ... & ... & ... & ... \\
				&&PB                                                 & -4.7(0) &0.36(\color{red}0.7)&  & ... & ... & ... & ... & ... & ... & ... & ... \\\cline{2-14}
				&\multirow{3}{*}{\texttt{LightGBM}}&ZIPB2 & 3.0(0.002) &13.3(0) &9.0(0)&  & ... & ... & ... & ... & ... & ... & ... \\
				&\multirow{3}{*}{}&ZIPB1                 &-1.3(\color{red}0.2)  &10.8(0)&4.7(0)& -6.4(0) &  & ... & ... & ... & ... & ... & ... \\
				&&PB                                                 &-1.3(\color{red}0.2)  &4.8(0)&6.0(0) &-5.4(0)&-0.7(\color{red}0.5)&  & ... & ... & ... & ... & ... \\\cline{2-14}
				&\multirow{3}{*}{\texttt{CatBoost}}&ZIPB2 &2.3(0.02)  &6.8(0)&6.6(0)&0.4(\color{red}0.7)&3.0(0.002)&3.6(0)&&...&...&...&...\\
				&\multirow{3}{*}{}	&ZIPB1                  &6.4(0) &11.4(0)&8.0(0)&4.1(0)&7.3(0)&5.8(0)&2.8(0.006)&  & ... & ... & ... \\
				&&PB                                          &-0.2(\color{red}0.8)  &5.6(0)&5.2(0)&-3.4(0)&0.7(\color{red}0.5)&1.4(\color{red}0.1)&-3.6(0)&-5.7(0)&& ... & ... \\\cline{2-14}
				&&ZIPG          &-12.8(0) & -12.3(0) &-12.3(0)&-17.0(0)&-14.3(0)&-15.1(0)&-14.3(0)&-15.8(0)&-15.5(0)&& ... \\ \cline{3-14}
				&&PG                                                 &-13.4(0) &-13.1(0) &-16.8(0)&-18.0(0)&-15.1(0)&-16.8(0)&-15.2(0)&-16.4(0)&-16.9(0)&-1.5(0)&\\
				\bottomrule
		\end{tabular}} 
		\begin{tablenotes}
			\small
			\item Numbers represent Vuong statistics, with p-values shown in parentheses.
		\end{tablenotes}
	\end{threeparttable}
\end{table}

Eleven models were trained using the data, and their performance is presented in Table \ref{tab:metric STD}. Of all the models, the \texttt{CatBoost} ZIPB1 model stood out with the highest Pseudo $R^2$ and the lowest mean Deviance.  As revealed in Table \ref{tab:metric freMTPL}, \texttt{CatBoost} models outperformed \texttt{LightGBM} and \texttt{XGBoost} in terms of ZIPB2, ZIPB1, and PB. All ZIPB2 models showed excellent performance following the \texttt{CatBoost} ZIPB1 model. Considering the Vuong statistics and its associated p-value presented in Table \ref{tab:vuong STD}, the models' performance is ranked as follows: \texttt{CatBoost} ZIPB1$>$\texttt{CatBoost} ZIPB2, \texttt{LightGBM} ZIPB2$>$\texttt{XGBoost} ZIPB2, \texttt{LightGBM} ZIPB1, \texttt{CatBoost} PB, \texttt{LightGBM} PB$>$ \texttt{XGBoost} ZIP1, \texttt{XGBoost} PB$>$ZIPG $>$PG.

\begin{figure}[htbp]
	\centering
	\includegraphics[scale=0.35]{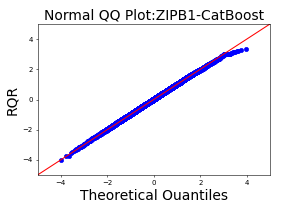}
		\includegraphics[scale=0.35]{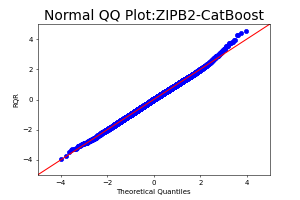}
			\includegraphics[scale=0.35]{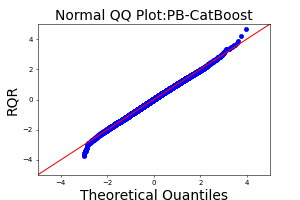}
				\includegraphics[scale=0.35]{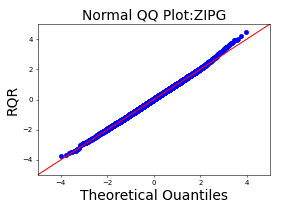}
					\includegraphics[scale=0.35]{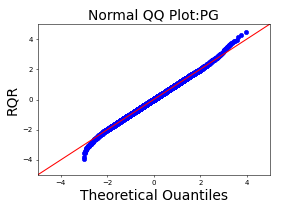}
	\caption{Q-Q Plots of Randomized Quantile Residuals: Five Models for Synthetic Telematics Data - \texttt{CatBoost} ZIPB1 (upper left), \texttt{CatBoost} ZIPB2 (upper middle), \texttt{CatBoost} PB (upper right), ZIPG (bottom left), PG (bottom right)}	\label{fig2:RQR_syn}
	\end{figure}

Similar to the result of Figure \ref{fig7:RQR_fre}, Figure \ref{fig2:RQR_syn} presents a Normal Q-Q plot of Randomized Quantile Residuals for five models fitted to the telematics data, demonstrating significant improvements in the tail of the estimated residuals of ZIPB1, ZIPB2, and ZIPG when compared with PB and PG. ZIPB1 and ZIPB2 appear to align samples with the standard normal distribution. This finding is consistent with the results presented in Table \ref{tab:metric STD} and \ref{tab:vuong STD}. Thus, the \texttt{CatBoost} zero-inflated Poisson boosted tree model, assuming that $p$ is a function of $\mu$, yields the best outcome, signifying that it is the most appropriate model for fitting and explaining the claim frequency of the synthetic telematics data. 

In conclusion, our analysis of the two datasets suggests that \texttt{CatBoost} is the most suitable library for developing auto claim frequency models. The ZIPB1 and ZIPB2 models have been demonstrated to be especially effective for zero-inflated claim data compared to other models. For the telematics data, the \texttt{CatBoost} ZIPB1 model outperforms the others. The French MTPL data indicates a preference for ZIPB2 over ZIPB1, however the latter still exhibits superior performance in comparison to other models. Moreover, the \texttt{CatBoost} ZIPB1 model provides a distinct advantage in allowing for the interpretation and analysis of risk features, leveraging various tools within the \texttt{CatBoost} library. Therefore, if the objective of modeling is to understand and interpret risk features, the ZIPB1 model is the most suitable choice.

In summary, based on the analysis conducted using the two datasets, it appears that \texttt{CatBoost} is more fitting than the other two libraries for developing auto claim frequency models. The ZIPB1 or ZIPB2 models have been found to be more appropriate for claim data relative to other models, relying on the data characteristics. As we mentioned before, when utilizing the \texttt{CatBoost} ZIPB1 model, it is advantageous to have the ability to interpret and analyze risk features in the model using various analysis tools within the \texttt{CatBoost} library.

\begin{figure}[htbp]
	\centering
	\includegraphics[scale=0.6]{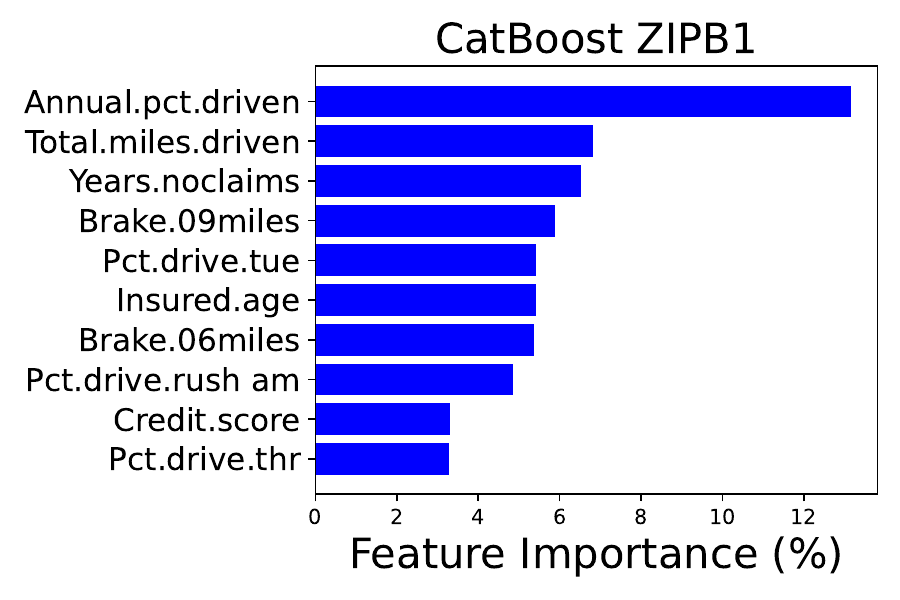}
	\caption{Top 10 Feature Importance in \texttt{CatBoost} ZIPB1}	\label{fig3:featureimp}
\end{figure}

\textbf{Interpretation.} Using \texttt{CatBoost} for the construction of an auto insurance frequency model presents numerous advantages. In particular, it includes various model analysis tools to interpret and evaluate the effects and interactions of different risk features on claim frequency. One primary method involves ranking features according to their impact on the change in predictions. This process is effectively visualized through a Feature Importance plot. The calculation of feature importance in \texttt{CatBoost} follows this formula:
\[\text{feature importance}_F = \sum_{\substack{\text{all trees}\\ \text{all leafs split by $F$}}}(v_1-m)^2 c_1+(v_2-m)^2 c_2.\]
Here, $m=\displaystyle\frac{v_1 c_1+v_2 c_2}{c_1+c_2}$. $c_1$ and $c_2$ denote the number of observations in the left and right leaves, respectively, while $v_1$ and $v_2$ represent the leaf values in the corresponding leaves. The feature importance values are normalized so that the sum of all feature importances equals 100. Figure \ref{fig3:featureimp} illustrates the Top 10 ranked feature importances for the ZIPB1 model, which was trained using the \texttt{CatBoost} library on synthetic telematics data.  The ``Annual.pct.driven" feature was ranked highest, followed by ``Total.miles.driven", ``Years.noclaims", and ``Brake.09miles". Of the top 10 features, only three are traditional, emphasizing the significant role telematics features play in determining claim frequency.

\begin{figure}[htbp]
	\centering
	\includegraphics[scale=0.6]{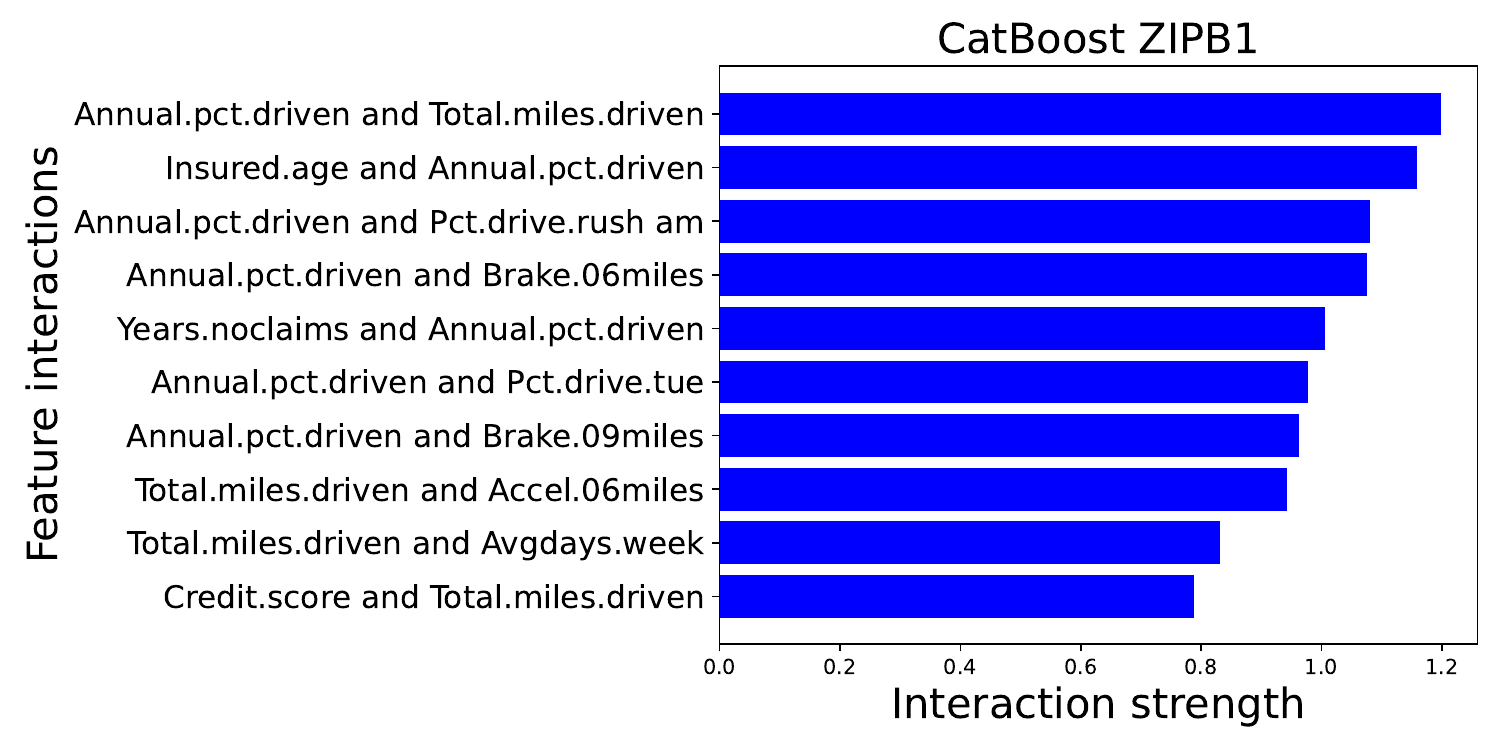}
	\caption{Top 10 Features Interation Strength in \texttt{CatBoost} ZIPB1}	\label{fig4:interaction}
\end{figure}

Apart from feature importance, which is common across GBDT models, \texttt{CatBoost} also provides the value of feature interaction strength. Within the \texttt{CatBoost} model, feature combinations are used as distinct features when splitting trees \cite{dorogush2018catboost}. This allows the model to measure interaction strength between features. All tree splits that include both features are observed to determine interaction. If splits of both features are present in the tree, then we are looking on how much leaf value changes when these splits have combined in the split and they are not combined. The calculation of interaction strength is as follows:
\[\text{interation }(F^1,F^2) =\sum_{{\substack{\text{all trees} \\ \text{with }F^1 \text{ and } F^2}}}\left| \sum_{F^1:F^2} \text{leaf value}-\sum_{F^1, \;F^2} \text{leaf value}\right|. \]
Here, the interaction strength between pairs of features $F^1$ and $F^2$ is quantified by the sum of absolute differences between the leaves of a tree that contains the $F^1:F^2$ combination and those that do not. Figure \ref{fig4:interaction} illustrates the top 10 feature interactions for the \texttt{CatBoost} ZIPB1 model on synthetic telematics data. The highest interaction strength was observed between the  ``Annual.pct.driven" and the ``Total.miles.driven". It is worth noting that these two features rank highly on the Feature Importance Plot, Figure  \ref{fig3:featureimp}. Additionally, the next four highest-ranked interactions involve the feature  ``Annual.pct.driven", paired with the ``Insured.age", ``Pct.drive.rush am", and ``Brake.06miles".

\begin{figure}[htbp]
	\centering
	\includegraphics[scale=0.6]{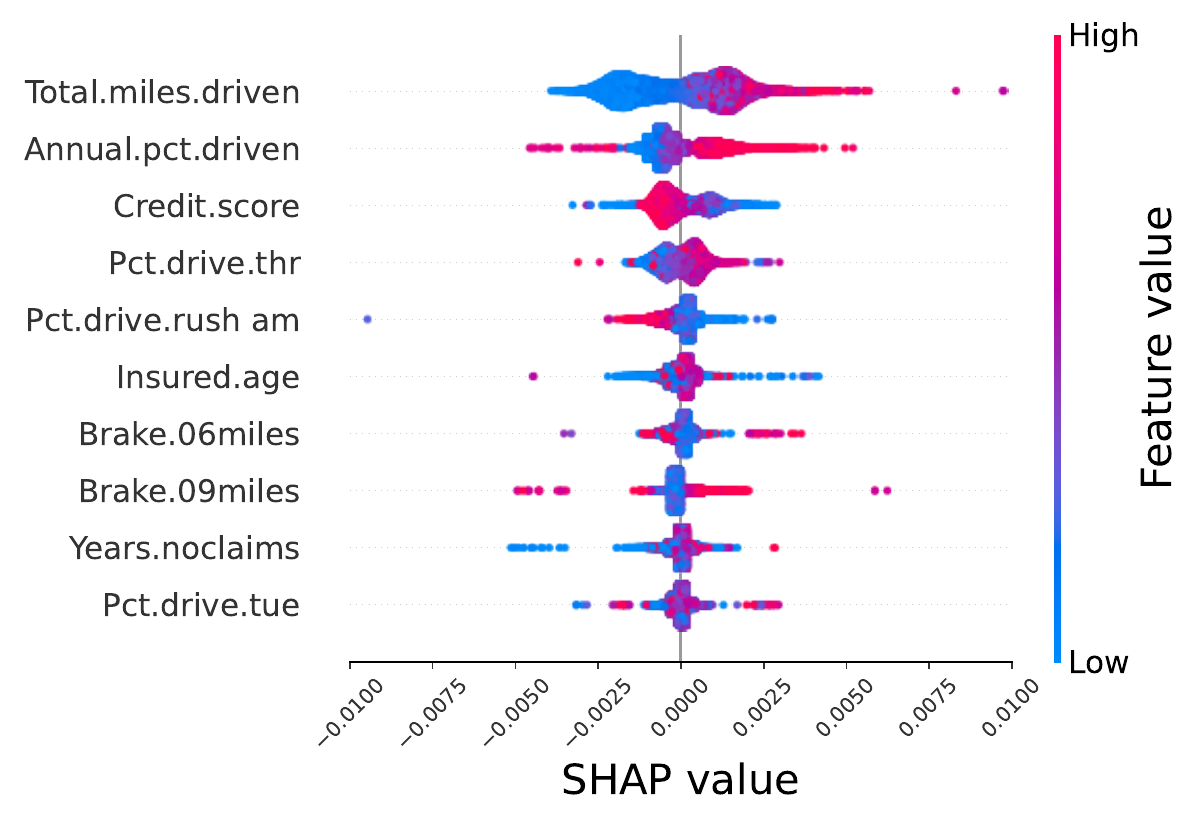}
	\caption{SHAP Values of Top 10 Features in \texttt{CatBoost} ZIPB1}	\label{fig5:shap}
\end{figure}

Like other libraries such as \texttt{XGBoost} and \texttt{LightGBM}, \texttt{CatBoost} is compatible with the SHAP Python library. This compatibility facilitates the use of SHAP values through the various analytic tools available in the SHAP package. SHAP (SHapley Additive exPlanations) is a game theory-based method that evaluates each feature's contribution towards the final prediction for a given observation \cite{lundberg2020local}. If a given observation's SHAP value for a specific feature is positive, it means that that feature's value contributes to an increase in the prediction value, taking it above the average prediction value. Figure \ref{fig5:shap} illustrates the SHAP values for the top 10 features, as outlined in Figure \ref{fig3:featureimp}. The key findings are:

\begin{itemize}
	\item High values of ``Total.miles.driven" increase claim frequency, while low values decrease it. 
	\item ``Annual.pct.driven" shows complex interaction effects, as high values can either increase or decrease claim frequency. This implies that the feature interacts with other features in determining claim frequency.
	\item Generally, a superior credit score leads to a reduction in claim frequency, although a lower credit score can induce either an increase or a decrease in claim frequency.
	\item  A higher percentage of driving on Thursdays correlates with increased claim frequency, while a lower percentage shows a reduction. This pattern might be indicative of weekly patterns in driver fatigue.
	\item The percentage of AM rush hour driving demonstrates that higher values correspond to decreased claim frequency, whereas lower values show an increase. This could suggest that drivers exercise more caution during their morning commute to work
	\item The insured age appears to have a complex effect; Middle values typically result in an increased claim frequency, high values tend to decrease it, while low values produce mixed results.
	\item The number of sudden brakes, defined as those exceeding 6 mph/s per 1000 miles, contributes to an increased risk at lower levels, but shows varied outcomes at higher levels.
	\item On the other hand, the number of sudden brakes exceeding 9 mph/s per 1000 miles tends to decrease risk at lower levels, while again showing mixed results at higher levels.
\end{itemize}

\begin{figure}[htbp]
	\centering
	\includegraphics[scale=0.4]{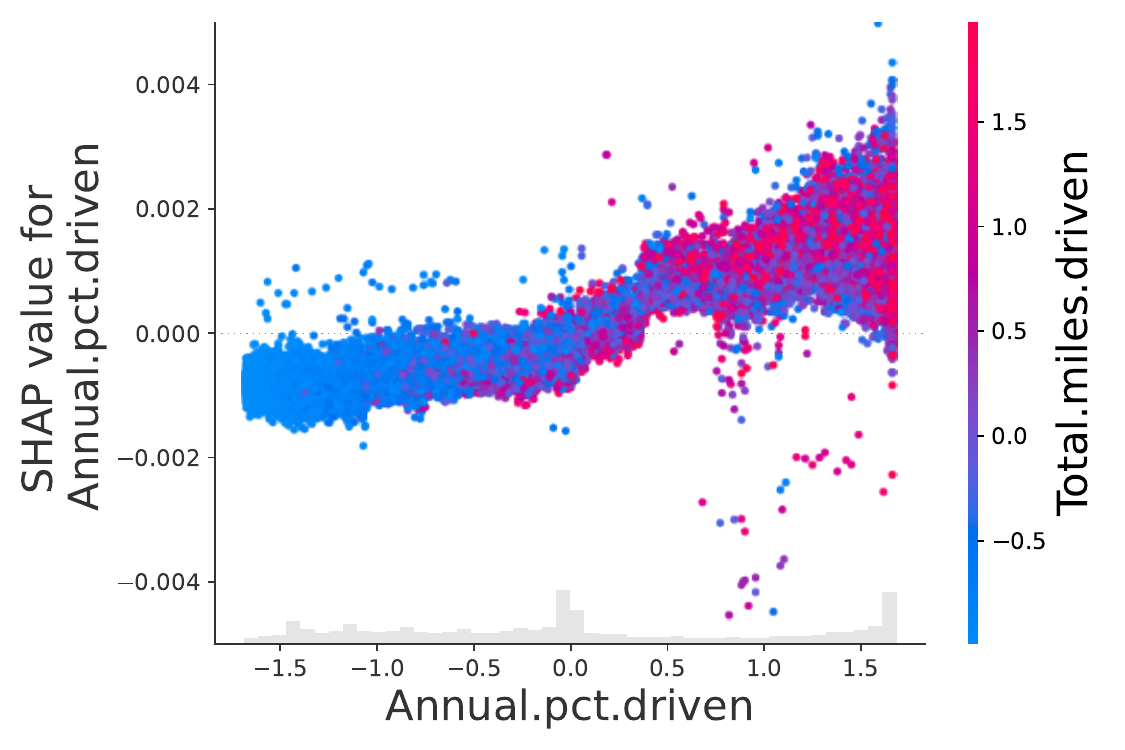}
	\includegraphics[scale=0.4]{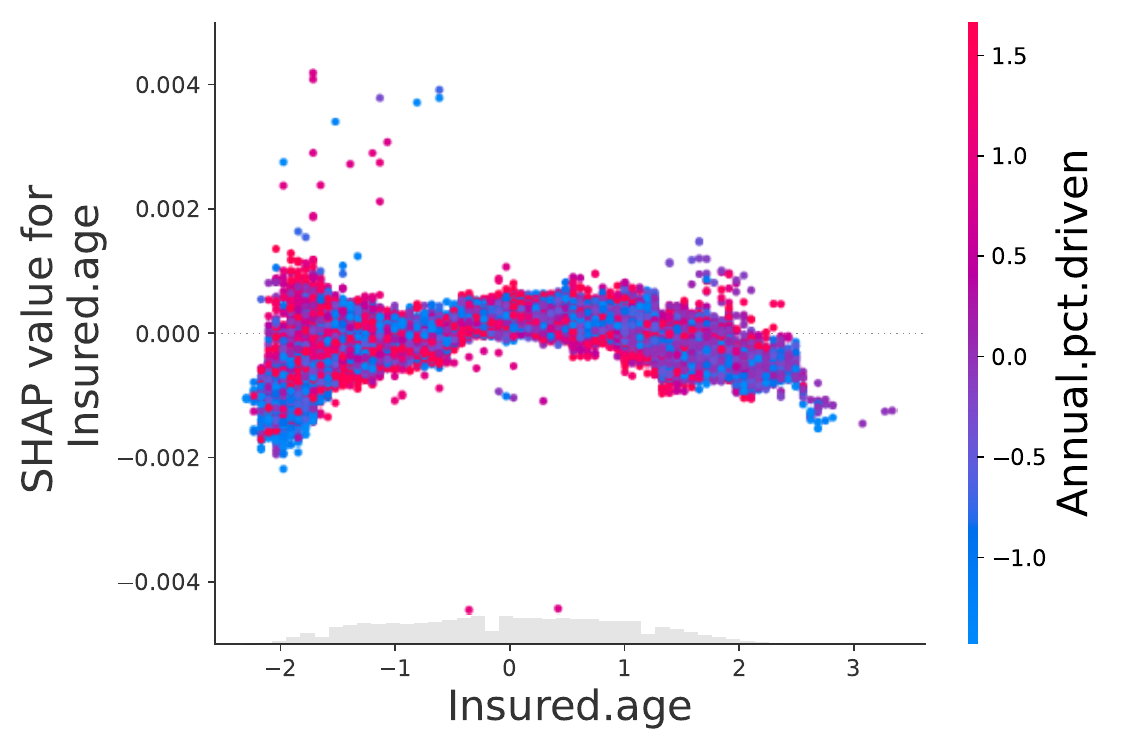}
	\includegraphics[scale=0.4]{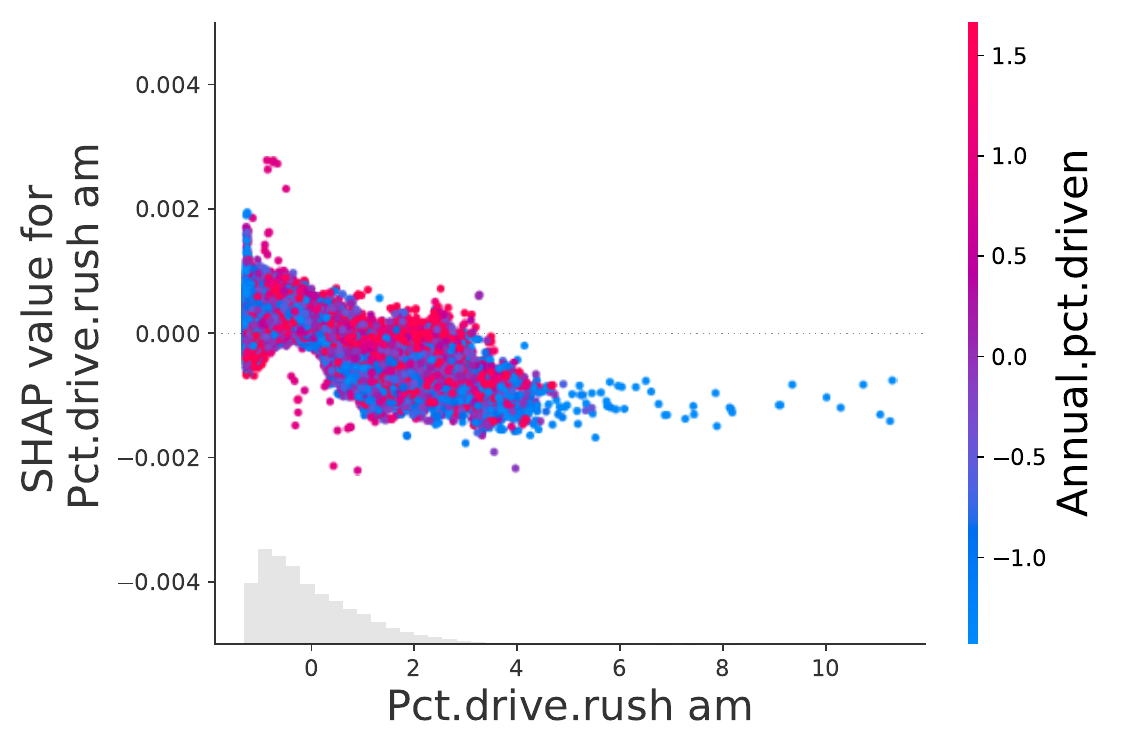}
	\includegraphics[scale=0.4]{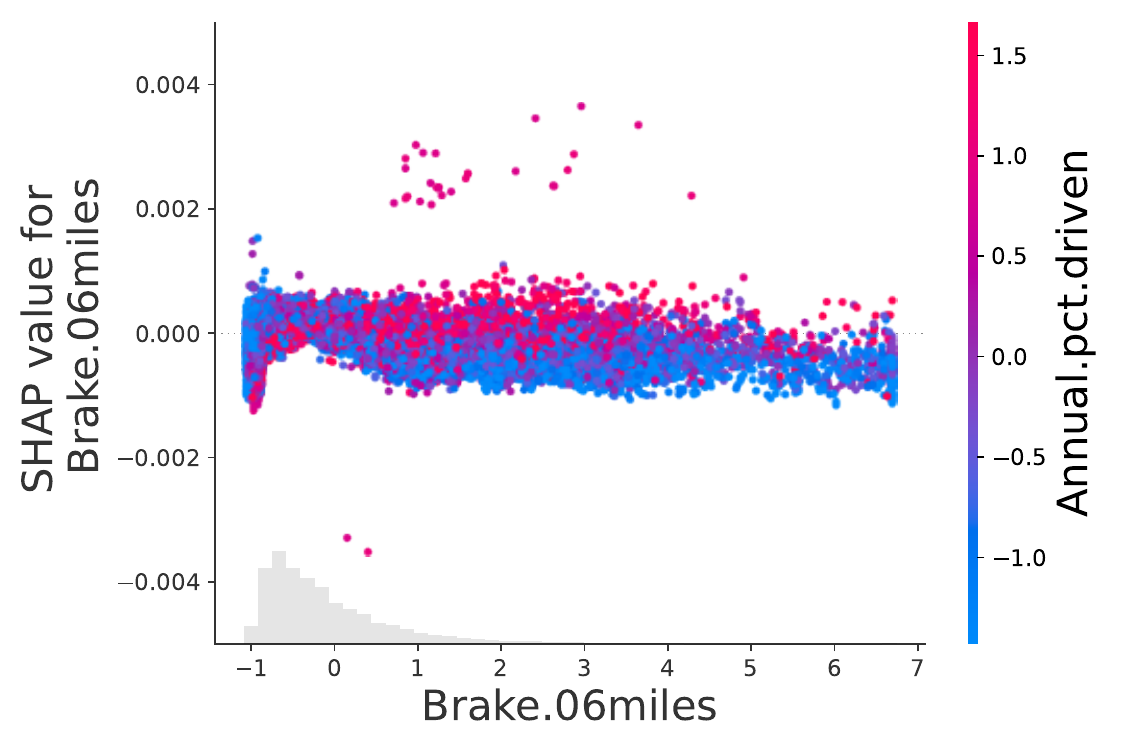}
	\caption{Four Scatter Plots Illustrating Feature Dependence Through SHAP Values}	\label{fig6:shapinteration}
\end{figure}

For a deeper examination of the interaction effects between features, we generated scatter plots of the SHAP values for each feature across all observations, with points color-coded by another feature. This is illustrated in  Figure \ref{fig6:shapinteration}. Depending on the feature interaction strength results shown in Figure \ref{fig4:interaction}, four plots were constructed, each representing the four most significant interdependencies among the features. The values on the x and y axes correspond to normalized values of each feature. 

From left top plot in Figure \ref{fig6:shapinteration}, we observed a positive correlation between the total annual miles driven and the annualized percentage of time spent on the road. For the insured age, it appears that low values can either increase or decrease claim frequency, depending on the value of ``Annual.Pct.Driven"; higher values increase claim frequency, while lower values decrease it. In the case of the percent of AM rush hour driving, we identified a decreasing trend. That is, an increase in this value tends to reduce claim frequency. However, it was observed that higher ``Annual.Pct.Driven" values intensify the risk more than lower values when the percentage of AM rush hour driving is high. The number of sudden brakes with a magnitude of 6 mph/s per 1000 miles, when low, appears to raise claim frequency. Interestingly, it was noted that when the value of `Brake.06miles'  is high, high values of `Annual.Pct.Driven' increase claim frequency while a decrease in claim frequency was associated with lower values.

\section{Concluding remarks} \label{sec:conclude}
Predicting insurance claim patterns in the property and casualty (P\&C) insurance industry is difficult due to the right-skewed distribution of claims, which is characterized by a large number of zero-valued, non-occurring claims. Traditional models, such as Poisson or negative binomial Generalized Linear Models (GLMs), have difficulty dealing with these excess zeros. To address this issue, ``zero-inflated" models have been developed, combining a traditional count model with a binary model to effectively manage such data sets.
Due to their remarkable predictive accuracy, gradient boosting techniques have become increasingly popular for creating auto insurance claim frequency models. 

This research adds to the existing literature by introducing and assessing a new boosting algorithm specifically designed for insurance claim data, particularly telematics data with a large number of zeros, to construct frequency models. We compare and evaluate popular gradient boosting libraries, \texttt{XGBoost}, \texttt{LightGBM}, and \texttt{CatBoost}, to determine which is best suited for processing insurance claim data and aligning with the proposed actuarial frequency models. After a thorough examination of two different datasets, we conclude that \texttt{CatBoost} is the optimal choice for auto claim frequency models, especially when dealing with heterogeneous datasets, as it produces the most accurate predictions. Depending on the data characteristics, zero-inflated Poisson Boosted Tree models — namely, ZIPB1, which assumes the inflation probability $p$ as a function of the distribution mean $\mu$, and ZIPB2, which treats $p$ and $\mu$ as independent — demonstrated superior performance relative to other models. An intrinsic advantage of the ZIPB1 model is its compatibility with specialized analytical tools within the \texttt{CatBoost} library, which enables a thorough examination of the effects and interactions of different risk factors on the frequency model, particularly in the context of telematics data. By using the \texttt{CatBoost} ZIPB1 model, we were able to identify and explain the interactions between telematics features and determine the influence of each key feature on the prediction of the claim amount.

Given these evident merits of ZIPB1, forthcoming research endeavors might be directed towards identifying the specific data attributes that underpin the pronounced efficacy of the ZIPB1 model over ZIPB2. Additionally, implementing a Negative Binomial Boosted tree for the NB regression model, with the assumptions suggested under both the ZIPB1 and ZIPB2 models, represents a valuable direction for future work. Unquestionably, this study's insights present a substantial augmentation to actuarial scholarship, more so in the domain of predictive modeling in insurance.

\clearpage
\subsection*{Appendix A. Detailed steps of ZIPB1 and ZIPB2 algorithms} \label{appa-alg}

\begin{algorithm}[H]
	\KwIn{Training dataset $D= \{\bm{x}_i , y_i \}_1^N$, total iterations $T$, learning rate $\alpha$, regularization parameter $\lambda$, inflation parameter $\gamma$}
	\KwOut{Final model: $\hat{\mu_i}=w_i \exp(F_T (\bm{x}_i))$, $\hat{p_i}=\displaystyle\frac{1}{1+\hat{\mu_i}^{\gamma}}$}
	Initialize tree:  $f_0 = 0$ \;
	\For{$t=1, \ldots, T$}{
		Compute the first(\ref{eq:18}) and second(\ref{eq:19}) derivatives $g_{i,t}=\partial_{F_{t-1}} L(y_i,F_{t-1}(\bm{x}_i))$ and $h_{i,t}=\partial^2_{F_{t-1}} L(y_i,F_{t-1}(\bm{x}_i))$ for each $i=1,\ldots, N$ \;
		Fit the tree $f_t$ minimizing loss function(\ref{eq:9})\;
		Update $\ F_t=F_{t-1}+\alpha f_t$\;
	}
	Return final prediction score: $\ F_T(\bm{x}_i) = \sum_{t=1}^{T} \alpha  f_t (\bm{x}_i) $ \;
	\caption{Zero-Inflated Poisson Boosted Tree Case 1: $p$ and $\mu$ Related (ZIPB1)}\label{alg:case1}
\end{algorithm}

\begin{algorithm}[H]
	\KwIn{Training dataset $D= \{\bm{x}_i , y_i \}_1^N$, total iterations $T$, learning rate $\alpha$, regularization parameter $\lambda$}
	\KwOut{Final model: $\hat{\mu_i}=w_i \exp(F_T^{po} (\bm{x}_i))$, $\hat{p}=\text{logit}^{-1}(F^{logit}_T(\bm{x}))$}
	Initialize trees:  $f^{po}_0 = 0$, $f^{logit}_0 = 0$ \;
	\For{$t=1, \ldots, T$}{
		Compute first(\ref{eq:21}) and second(\ref{eq:22}) derivatives $g^{po}_{i,t}=\partial_{F^{po}_{t-1}} L(y_i,F^{po}_{t-1}(\bm{x}_i),F^{logit}_{t-1}(\bm{x}_i))$ and $h^{po}_{i,t}=\partial^2_{F^{po}_{t-1}} L(y_i,F^{po}_{t-1}(\bm{x}_i),F^{logit}_{t-1}(\bm{x}_i))$ for each $i=1,\ldots, N$ \;
		Fit the tree $f^{po}_t$ minimizing loss function(\ref{eq:9})\;
		Update $\ F^{po}_t=F^{po}_{t-1}+\alpha f^{po}_t$\;
		
		Compute first(\ref{eq:23}) and second(\ref{eq:24}) derivatives $g^{logit}_{i,t}=\partial_{F^{logit}_{t-1}} L(y_i,F^{po}_{t}(\bm{x}_i),F^{logit}_{t-1}(\bm{x}_i))$ and $h^{logit}_{i,t}=\partial^2_{F^{logit}_{t-1}} L(y_i,F^{po}_{t}(\bm{x}_i),F^{logit}_{t-1}(\bm{x}_i))$ for each $i=1,\ldots, N$ \;
		Fit the tree $f^{logit}_t$ minimizing loss function(\ref{eq:9})\;
		Update $\ F^{logit}_t=F^{logit}_{t-1}+\alpha f^{logit}_t$\;
		
	}
	Return final prediction scores: $\ F^{po}_T(\bm{x}_i) = \sum_{t=1}^{T} \alpha  f^{po}_t (\bm{x}_i) $ and $\ F^{logit}_T(\bm{x}_i) = \sum_{t=1}^{T} \alpha  f^{logit}_t (\bm{x}_i) $\;
	\caption{Zero-Inflated Poisson Boosted Tree Case 2: $p$ and $\mu$ Unrelated (ZIPB2)}\label{alg:case2}
\end{algorithm}

\clearpage
\subsection*{Appendix B. Descriptive details of datasets} \label{appb-variables}

\begin{table}[H]
	\centering
	\caption{Variable Names and Descriptions for the French Motor Third-Party Liability} \label{tab:VD1}
\begin{threeparttable}
	\resizebox{!}{3cm}{
		\begin{tabular}{lll}
			\\
			\toprule
			Type & Variable  & Description \\
			\midrule
			Traditional & Exposure & Total exposure in yearly units \\
			&  Area** &Area code: A,B,C,D,E\\
			& VehPower & Power of the car \\
			&  VehAge& Age of the car in years\\
			&  DrivAge& Age of the main driver in years\\
			& BonusMalus & Bonus-malus level between 50 and 230 (with reference level 100)\\
			& VehBrand** & Car brand: 11 lables in \{B1-B6, B10-B14 \}\\
			& VehGas**& Vehicle fuel type: Diesel, Regular fuel \\
			&  Density& Density of inhabitants per $\text{km}^2$ in the city of the living place of the driver      \\
			&  Region**& Regions in France: 22 lables in \{R11, R21, R22, R24,$\cdots$, R93\} \\
			\midrule
			Response & ClaimNb& Number of claims on the given policy \\
			\bottomrule
	\end{tabular}}
 \begin{tablenotes}
	\small
	\item [**] Indicates categorical variable.
\end{tablenotes}
\end{threeparttable}
\end{table}

\begin{table}[H]
	\centering
	\caption{Variable Names and Descriptions for the Synthetic Telematics Dataset} \label{tab:VD2}
	\begin{threeparttable}
	\resizebox{!}{5.9cm}{
		\begin{tabular}{lll}
			\\
			\toprule
			Type & Variable  & Description \\
			\midrule
			Traditional & Duration & Total exposure in yearly units \\ 
			& Insured.age  & Age of insured driver \\
			& Insured.sex**  & Sex of insured driver: Male, Female \\
			& Car.age  & Age of vehicle (in years) \\
			& Marital**  & Marital status: Single, Married \\
			& Car.use**  & Use of vehicle: Private, Commute, Farmer, Commercial  \\
			& Credit.score  & Credit score of insured driver \\
			& Region**  & Type of region where driver lives: Rural, Urban \\
			& Annual.miles.drive  & Annual miles expected to be driven declared by driver \\
			& Years.noclaims  & Number of years without any claims\\
			& Territory**  & Territorial location of vehicle: 55 labels in \{11, 12, 13, . . ., 91\}\\
			\midrule
			Telematics & Annual.pct.driven  & Annualized percentage of time on the road \\
			& Total.miles.driven  &Total distance driven in miles \\
			&Pct.drive.xxx	&Percent of driving day xxx of the week: mon/tue/…/sun\\
			&Pct.drive.x hrs&Percent vehicle driven within x hrs: 2hrs/3hrs/4hrs\\
			&Pct.drive.xxx	&Percent vehicle driven during xxx: wkday/wkend\\
			&Pct.drive.rush xx	&Percent of driving during xx rush hours: am/pm\\
			&Avgdays.week	&Mean number of days used per week\\
			&Accel.xxmiles	&Number of sudden acceleration 6/8/9/…/14 mph/s per 1000miles\\
			&Brake.xxmiles	&Number of sudden brakes 6/8/9/…/14 mph/s per 1000miles\\
			&Left.turn.intensityxx	&Number of left turn per 1000miles with intensity xx: 08/09/10/11/12\\
			&Right.turn.intensityxx	&Number of right turn per 1000miles with intensity xx: 08/09/10/11/12\\
			\midrule
			Response & NB\_Claim  & Number of claims on the given policy \\
			& AMT\_Claim  & Amount of claims on the given policy \\
			\bottomrule
	\end{tabular}}
 \begin{tablenotes}
	\small
	\item [**] Indicates categorical variable.
\end{tablenotes}
\end{threeparttable}
\end{table}

\bigskip

\bibliographystyle{apalike}
\bibliography{catboost.bib}

\end{document}